  \documentclass[letterpaper, 10pt, twocolumn]{article}
  \usepackage{clean}
  \shorttitle{Compliant gripper for assembly of electrical components}
\usepackage{gensymb}
\usepackage{adjustbox}
\usepackage{graphicx}
\usepackage{subcaption}
\graphicspath{{./figs}}
\usepackage{epsfig}

\usepackage{amsmath,amssymb,amsfonts}
\usepackage{algorithm2e}
\usepackage{xcolor}
\usepackage[colorlinks=true, linkcolor=black, urlcolor=cyan, filecolor=black, citecolor=black]{hyperref}
\usepackage{url}
\usepackage{amsmath,bm}
\usepackage{multirow, makecell} % For table magic
\usepackage{adjustbox}
\usepackage{comment}

\begin{document}

%%% Possible journal submissions
% RA-L Submission, <6 pages https://www.ieee-ras.org/publications/ra-l/submission-procedures
% IEEE T-ASE Submission, <10 Pages https://www.ieee-ras.org/publications/t-ase/information-for-authors-t-ase
% RCIM, no page limit found https://www.sciencedirect.com/journal/robotics-and-computer-integrated-manufacturing

\title{Compliant finray-effect gripper for high-speed robotic assembly of electrical components}
\author{Richard Matthias Hartisch$^1$ and \hspace{0.03em} Kevin Haninger$^2$ 
\thanks{$^1$ Department of Industrial Automation Technology at TU Berlin, Germany. \\ $^2$ Department of Automation at Fraunhofer IPK, Berlin, Germany.  \\ Corresponding author: {\tt r.hartisch@tu-berlin.de}}
\thanks{This project has received funding from the European Union's Horizon 2020 research and innovation programme under grant agreement No 820689 — SHERLOCK and 101058521 — CONVERGING.}}

\maketitle

\begin{abstract}

Fine assembly tasks such as electrical connector insertion have tight tolerances and sensitive components, limiting the speed and robustness of robot assembly, even when using vision, tactile, or force sensors. Connector insertion is a common industrial task, requiring horizontal alignment errors to be compensated with minimal force, then sufficient force to be brought in the insertion direction. The ability to handle a variety of objects, achieve high-speeds, and handle a wide range in object position variation are also desired. Soft grippers can allow the gripping of parts with variation in surface geometry, but often focus on gripping alone and may not be able to bring the assembly forces required. To achieve high-speed connector insertion, this paper proposes monolithic fingers with structured compliance and form-closure features. A finray-effect gripper is adapted to realize structured (i.e. directional) stiffness that allows high-speed mechanical search, self-alignment in insertion, and sufficient assembly force. The design of the finray ribs and fingertips are investigated, with a final design allowing plug insertion with a tolerance window of up to 7.5 mm at high speed.  

\end{abstract}

\section{Introduction}
\label{sec:intro}
Installation of cables and wire harnesses is increasingly important, especially as electrification of automobiles and household appliances increases. While the pre-production of cable harnesses (cutting, mounting of wire seals and attachment of cable heads) can be achieved with specialized machinery \cite{trommnau2019overviewneu}, installation is still largely manual work \cite{Yumbla.2020}. 

Cable installation is challenging to automate due to the high variety in connectors \cite{Yumbla.2020} which can lead to small batch sizes \cite{trommnau2019overviewneu}. The handling of cables also introduces technical challenges. Cable routing requires methods for deformable linear objects (DLOs) \cite{trommnau2019overviewneu, Chen.2016}. Many of the installation steps, e.g. connector mating, are fine assembly tasks which require coordinated vision and touch when done by humans \cite{Chen.2016}.
% \begin{figure}[t]
% \centering
%   \begin{subfigure}{0.7\linewidth}
%     \includegraphics[width=\columnwidth]{figs/KS_search2.png}
%     \caption{}\label{subfig:key-a}
%   \end{subfigure} \\ 
%     \begin{subfigure}{0.40\linewidth}
%       \includegraphics[width=\linewidth]{figs/Search strategy 2.jpg}
%       \caption{}\label{subfig:key-b}
%     \end{subfigure}
%     \qquad
%     \begin{subfigure}{0.40\linewidth}
%       \includegraphics[width=\linewidth]{figs/Search strategy 3.jpg}
%       \caption{}\label{subfig:key-c}
%     \end{subfigure}
    
%   \caption{Working principle of finray-effect gripper is demonstrated with varying types of plugs. In (a) the initial contact and first search direction in x, in (b) the alignment in y, in (c) the insertion is completed.}
%   \label{fig:example-key}
% \end{figure}
% \iffalse
\begin{figure}[t!]
  \begin{subfigure}{0.62\linewidth}
    \includegraphics[width=\columnwidth]{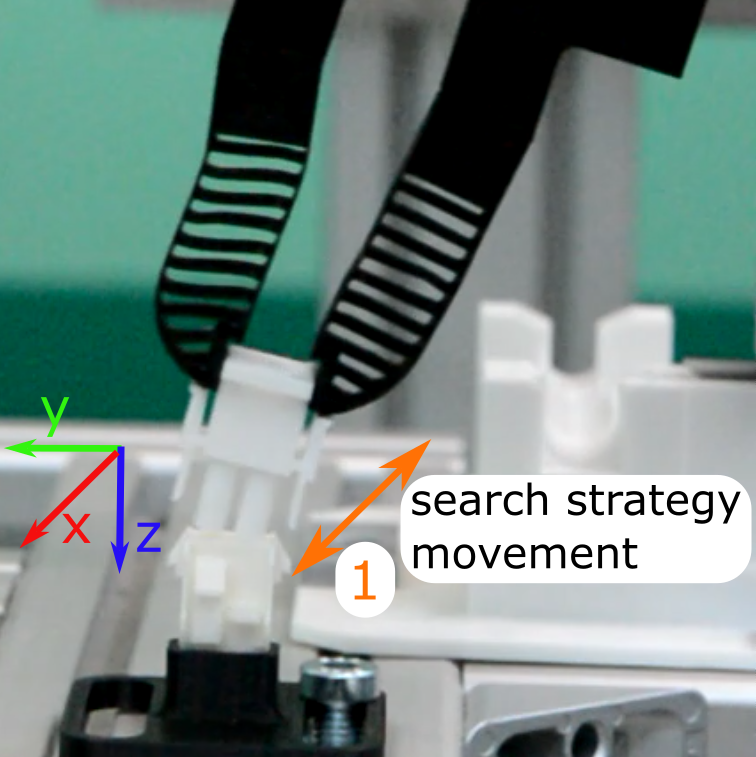}
    \caption{}\label{subfig:key-a}
  \end{subfigure}\hfill
  \begin{minipage}{0.35\linewidth}
    \begin{subfigure}{\linewidth}
      \includegraphics[width=\linewidth]{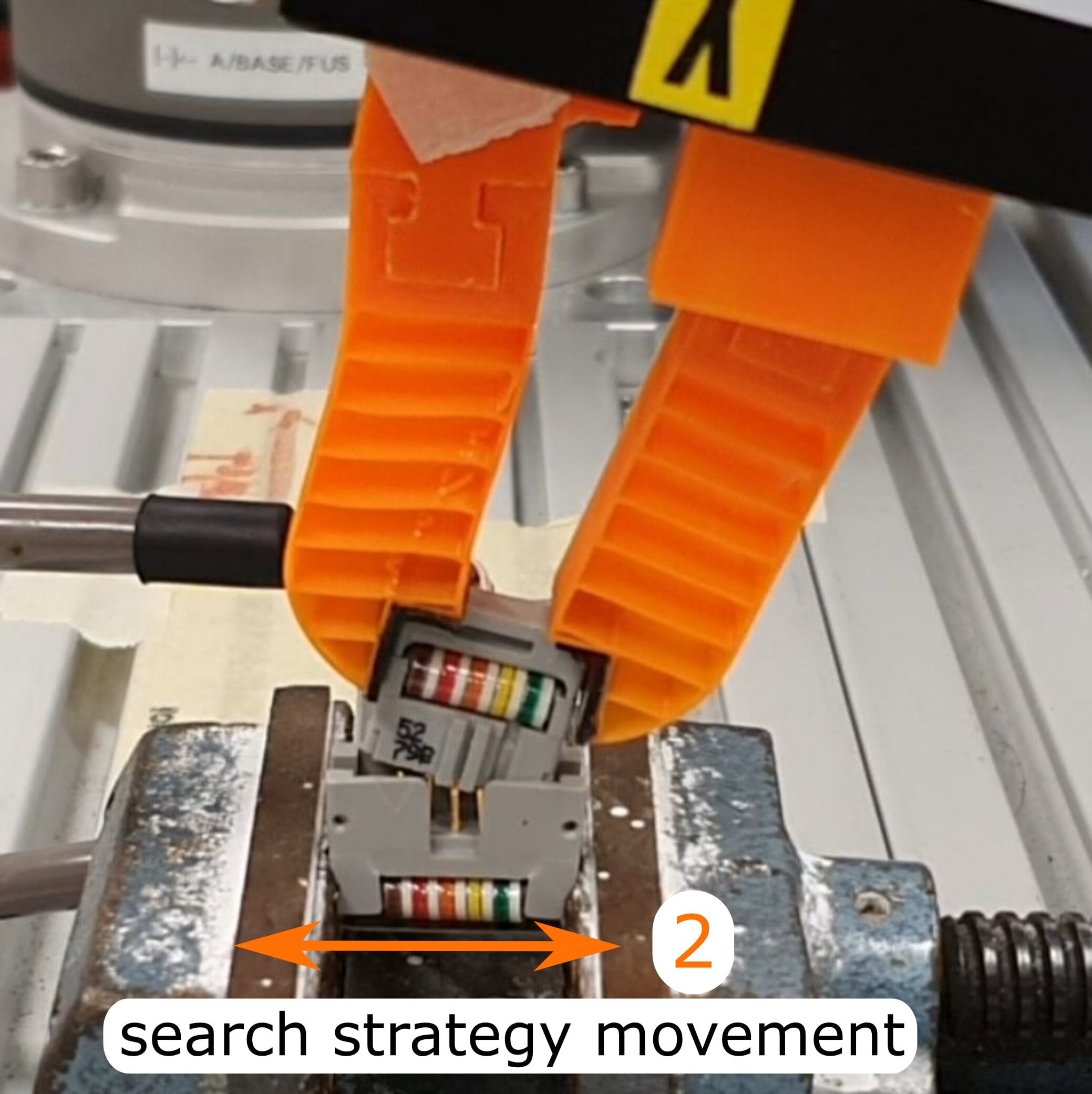}
      \caption{}\label{subfig:key-b}
    \end{subfigure}\hfill
    \medskip
    \begin{subfigure}{\linewidth}
      \includegraphics[width=\linewidth]{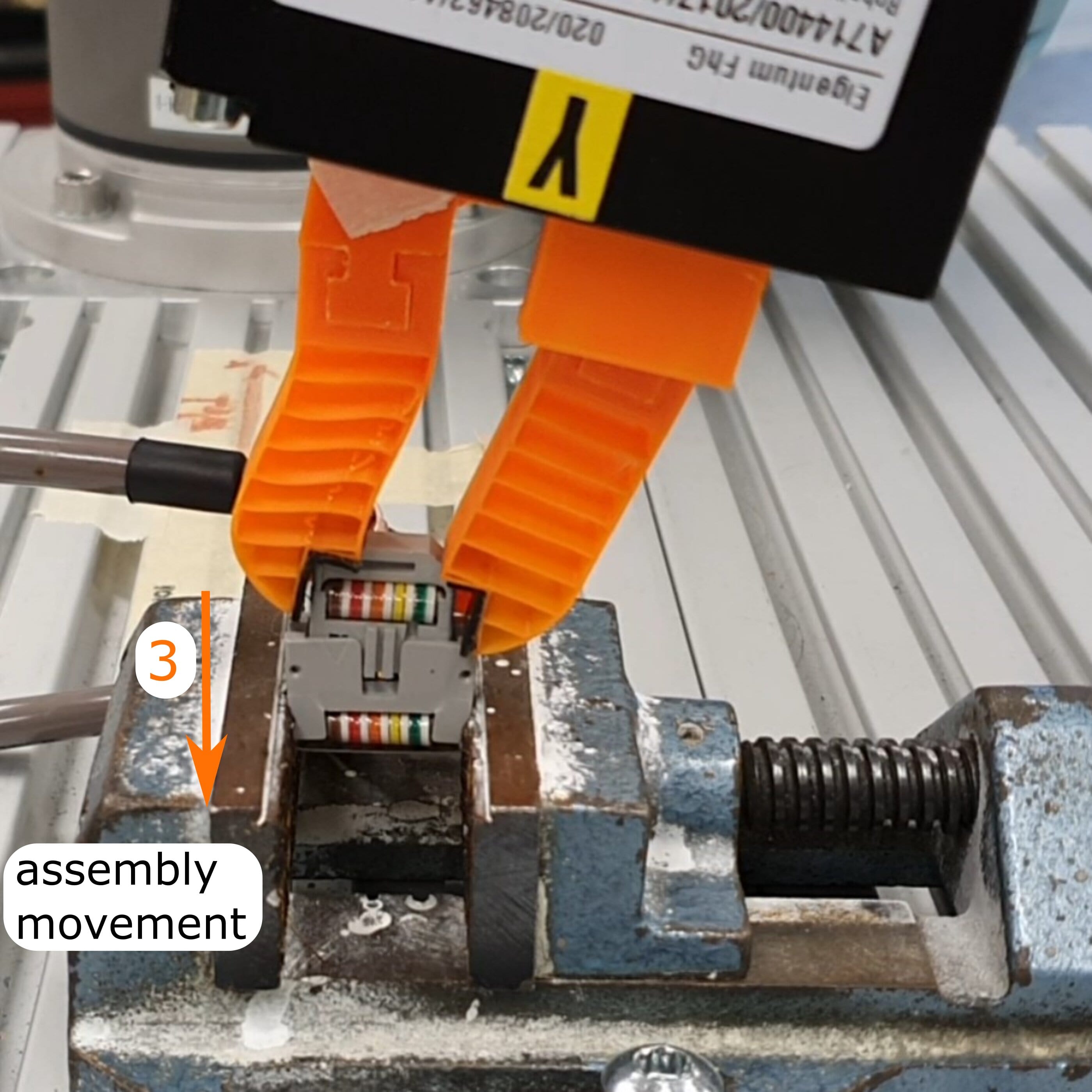}
      \caption{}\label{subfig:key-c}
    \end{subfigure}\hfill
  \end{minipage}
  \caption{Working principle of finray-effect gripper is demonstrated with varying types of plugs}
  \label{fig:example-key}
\end{figure}
% \fi

The mating of connectors can be divided into three steps: gripping, search, and insertion. By the end of the insertion, a certain relative pose between gripped part and target must be achieved. To be practical, this must be achieved over certain variation in target pose. An unknown or uncertain pose of the grasped plug inside the gripper is also a major contribution to the complexity \cite{li2014localization}. Complexity of search and insertion increases due to the tolerances between plug and socket, small parts, variation in plug geometry, limited grasping and contact area on the plug, limited free space near sockets, and the necessity of a high assembly force. Angular displacements can be tolerated to a certain degree, exceeding this results in a failed assembly \cite{Yumbla.2019}.

% \begin{figure}[t!] 
    
% 	\centering
% 	% \subfloat[Position of Clamp in Grip
% 	\label{clamp_pos}{\includegraphics[width=0.50\columnwidth]{figs/Fourth_Contact.jpg}}
% 	\hfill
% 	% \subfloat[Assembled Clamp Joint
% 	% \label{clamp_assmbl}]{\includegraphics[width=.40\columnwidth]{img/assembled clamp.jpg}}
% 	\protect\caption{Working Principle of finray-effect Gripper}
% 	%\label{ribbon}
% \end{figure}

% \begin{figure}[t]
%     \centering
%     \subfloat[
% 	\label{gripper1}]{\includegraphics[width=0.48\columnwidth]{figs/Gripper_cut.png}}
% 	\hfill
%     \subfloat[
% 	\label{gripper2}]{\includegraphics[width=.35\columnwidth]{figs/Fourth_Contact.jpg}}
%     \hfill
%     \subfloat[
% 	\label{gripper3}]{\includegraphics[width=.35\columnwidth]{figs/Fifth_Contact.jpg}}
%     \caption{Working principle of finray-effect Gripper is demonstrated with varying types of plugs}
%     \label{principle}
% \end{figure}

These challenges can be partly handled by compliance. In gripping, compliance can allow good contact area over some variation in the plug geometry. In the search and insertion process, compliance can compensate misalignment in the relative pose between gripped part and socket.   

Compliance can be divided into two parts: active and passive compliance \cite{wang1998passive}, where passive compliance is the intrinsic mechanical compliance of the physical structure, and active compliance is achieved by feedback controller design. A relevant example for compliance is the remote center of compliance (RCC) \cite{ciblak2003designneu, whitney2004mechanicalneu}, which allows self-alignment in insertion tasks. Active compliance, such as impedance or admittance control can adapt the relative pose between the mating parts automatically depending on position and forces during contact \cite{baksys2017vibratory}. 

The major advantage of active compliance is the possibility to digitally change compliance, e.g. adjusting the RCC location to improve performance \cite{wang1998passive}.  The disadvantages of active compliance are the relatively high costs and the limited bandwidth \cite{wang1998passive}, which typically leads to higher collision forces. In contrast, passive compliance has no bandwidth limits and can significantly reduce collision forces. However, passive compliance is harder to design and is usually determined iteratively, since analytical models are not common, resulting in a higher effort in the design work, experiments and parameter identification. Additionally, passive compliance is usually specified to a certain task or part, which limits the generalizability \cite{chen2017improved}.

Here, passive compliance is used to allow stable high-speed contact transitions. By using prototype-friendly fused deposition modeling, low-cost monolithic solutions can be provided which realize compliance through elastic deformation. Additionally, this allows quick and easy testing, accelerating the process of finding a suitable passive compliance.

This work's contribution is a novel structured compliance finger design for gripping and assembly of electrical components. These fingers improve the speed and robustness of position-controlled robots in such tasks. Compared with sensorized fingers for plug insertion \cite{wang2021, jiang2022}, the proposed compliance allows a larger tolerance window and higher speed. Compared with existing work on cable harnesses \cite{chen2012design, Yumbla.2019, trommnau2019overviewneu}, which provide a general overview of wire harness design and production, we provide a taxonomy and detailed requirements of electrical plug assembly and an analysis of design requirements. 
% \cite{trommnau2019overviewneu} provides a general overview of wire harness design and production processes, whereas we provide detailed requirements for electrical connector assembly. 
Compared with existing plug insertion approaches using active compliance, which take up to $\approx 6-16$s \cite{park2013intuitive}, the passive compliance allows a successful assembly of the connectors in $\approx 1.2$ s from first contact. 

The rest of the paper is organized as follows. Section \ref{sec:problem_desc} categorizes the parameters of the plugs used in this work, the parameters occurring in the assembly task and describes the steps of the assembly process. Section \ref{sec:design} introduces the final gripper design, derived from the finray-effect, where the design parameters, the design and manufacturing process and problems are described. A range of applications to verify the gripper's abilities are presented in section \ref{sec:app}, consisting of repeatability and robustness experiments to determine design parameters which achieve the widest tolerable scale of misalignment.
% and an exemplary assembly of clamp connections to demonstrate generalizability. 
Finally, the conclusion and future work is given in section \ref{sec:discussion}.

\section{Electrical connector problem description}
\label{sec:problem_desc}

This section analyzes the problem of connector assembly, providing a taxonomy of electrical connectors and the assembly process itself.

\subsection{Taxonomy of connectors}

\begin{table*}[ht]
\renewcommand{\arraystretch}{1.15} %<- modify value to suit your needs
\begin{center}
\caption{Identified important properties of connectors and the assembly task, what parts of the robotic solution they influence, and possible values that the property can take \label{tab:conn_parameters}}
\begin{tabular}{r r|l|l} 
& Property & Effects & Possible values \\
\hline
\multirow{5}{*}{\rotatebox[origin=c]{90}{\large Connector}} & Fit and tolerances & Search strategy, req'd assembly force & Press, running, transition \\
& Plug exposed after insert & Grip location in insertion & Flush, $>0$ mm \\
 & Cable gland orientation & Grip location and free space & Straight, right angle \\
& Pin height & Search strategy & Flush, $<0$ mm \\
& Securing feature & Insertion, validation & Clip, lever  \\
\hline
\multirow{5}{*}{\rotatebox[origin=c]{90}{\large Task}} & Plug availability & Grasp strategy, finger design & magazine, on table, cluttered \\
& Socket availability & Tolerances, search strategy & fixed position, in workpiece, free space \\
& Space requirements & Finger dimensions, robot strategy & free space dimensions \\
& Cable handling & Additional tasks & need insert clips, need to pull cable \\
& Validation & Insert strategy & Is a validation (e.g. push-pull-push) required? \\
\hline

\end{tabular}
\end{center}
\end{table*}
While there is large variation in connector design, several parameters have a substantial influence on the robotic solution, summarized in Table \ref{tab:conn_parameters}. These parameters can have an influence on the allowed finger design, as well as the strategy for grasping, searching, or insertion. 

Some parameters are shown in Figure \ref{fig:plug_grip}, left, which shows an inserted plug. The amount that the cable head sticks out of the socket determines how the cable has to be gripped. Some plugs are also flush, so after insertion no part of the plug remains exposed. The cable gland can have different orientations, either straight out or in a right angle into the plug, which changes what space must be left free by the finger design. The cable type can be categorized as either a ribbon cable, or single-/multi-cable.  Further, the pin height inside the plug and/or socket heavily influences the search pattern, as this could result in collision and jamming. However, the number of pins don't heavily influence the assembly process. Additional safety features, such as levers or clips, may require additional assembly force or post-processing to secure. The tolerances between the plug and socket influence the search strategy and required assembly force.

\subsection{Categorization of assembly task}

There are additional parameters in typical connector assembly tasks which affect the design. How the plug is supplied affects the uncertainty in grip pose, as the plug could either be fixed rigidly, e.g. in a magazine, lying freely on the table or placed in a cluttered environment. Similarly, the socket could be either be in a fixed position, integrated in a workpiece or also in a magazine. Further, space limits from the environment have to be regarded due to the finger dimensions and the space requirements from the robot during the search strategy. During the assembly additional cable and wire handling has to be considered, meaning if intermediate clips are necessary, a cable straightening is required etc. After successfully mating the components additional testing could be necessary, e.g. push-pull-push of the cable. 

\subsection{Grip, search, and insert strategies}
\label{subsec:strategies}
The complete assembly process is considered in three stages: grip, search, and insert. 

From an initial position, the plug is gripped. Without a magazine or jig providing a constant and known pose of the plug, a known pose or at least the orientation of the plug inside the grip should be established to certain tolerances in order to achieve a successful alignment between plug and socket. 
\begin{figure}[t]
\begin{center}
\includegraphics[width=0.32\textwidth]{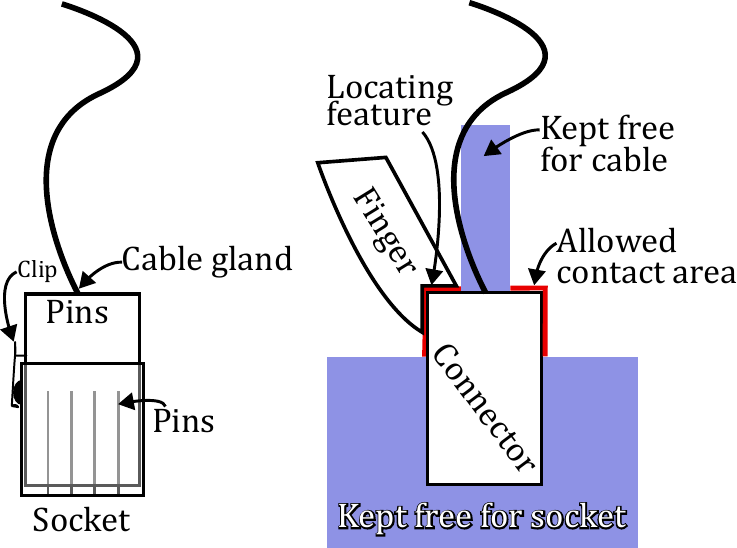}
\end{center}
\caption{Left, inserted connector in socket with key features, right, finger grasping the connector \label{fig:plug_grip}}
\end{figure}

The grasping strategy includes these aspects of the finger design, which are summarized in the right of Fig. \ref{fig:plug_grip}: (i) what contact area between the plug and finger can be used, (ii) what space around the connector must be kept free, (iii) are locating features needed to provide either repeatable position or sufficient assembly force? In addition to the fingertip design, the grasping strategy may include (i) a magazine for providing the plug in a semi-repeatable way and (ii) an adjustment strategy to ensure the connector is in a repeatable position in the fingers.  

The search strategy should achieve alignment of the plug and socket. When using mechanical search, the search strategy should be designed considering: (i) the variation in pose that needs to be covered with the strategy, (ii) the initial contact between plug and socket, which can be a point, line or planar contact depending on how the plug is presented, (iii) the height of the pins, which could be bent if contacted by the tip of the plug, (iv) validating that the plug has successfully slipped into the socket after the alignment.  

The mechanical search strategy here is shown in Fig. \ref{fig:example-key}.  Initial contact is made with a tilted plug, such that the corner of the plug lightly presses on the edge of the connector, Fig. \ref{subfig:key-a}. A motion in the x-direction allows the plug to slip into the socket when aligned. Next, contact is established with the sides of plug and socket, realized by a motion in the y-direction, Fig. \ref{subfig:key-b}. At this point, the leading corner of the plug should be slightly inserted and resting on the edge of the socket.

For the insertion phase following aspects should be considered: (i) the finger design should be able to avoid jamming of the connectors. For this, compliance, either active or passive, could be suitable, where the plug is able to rotate inside the socket due to the contact. (ii) The assembly force should not exceed a certain threshold, to avoid damaging the parts, which can be realized by the finger compliance.

% To mount the plug into the socket, the gripper moves in the z-direction, where the plug is able to align itself inside the socket due to the contact. 
% % Due to the passive compliance, the plug is able to rotate inside the socket from the contact. 
% After assembly, the robot rotates about the y-axis to avoid collision with the socket when opening the gripper and the plug is released. 

% * Grip plug from initial position
% * (possibly) Establish known pose of plug in grip
% * (possibly) Establish pose of socket
% * Approach socket
% * Align plug to socket
% ** Establish line contact with corner of plug with (light) positive contact force
% ** Search strategy
% ** X-motion over tolerance window (back and forth), plug should slip into socket in motion
% ** Y-motion to make contact between side of plug and socket, slight rotation about X
% ** (possibly) Validation that plug is slipped into socket
% ** Z-motion to mount, plug rotates from socket contact
% ** Robot rotates to avoid collision with socket when opening gripper
% * (possibly) Validate successful insertion
% * Release plug

\section{Design of Compliant Finray-effect Grippers}
\label{sec:design}

In this section, we describe the modification and parameterization of a finray-effect gripper \cite{crooks2016fin} to realize structured passive compliance. Where classical finray-effect grippers allow deformation to adapt to variation in surface geometry of grasped parts, we would like to find a design where the rendered compliance on a gripped part can be adjusted. 
ng features or notches into the fingertip, which would be necessary to achieve a high stiffness value in the assembly direction. 

Regarding the requirements and constraints in Sec. \ref{sec:problem_desc}, a monolithic compliant gripper can be built to successfully work in narrow environments. The finray design is used, and the flexibility of 3D printing is used to make the following modifications, seen in Figure \ref{finger_design_rotated}, to make it better suited to the problem:
\begin{itemize}
    \item Fingertip design: Necessary to achieve form fit and increase range of motion at fingertip
    \item Rib angle/ Infill direction
    \item Rib density/ Infill density
    \item Finger mounting angle
\end{itemize}

\subsection{Revised finray Design}

The finray-effect gripper mimics the deformation of fish fins, which are composed by two outer walls forming a V shape. Between the bones crossbeams are placed which determine the mechanical properties of the finray-effect gripper. 
The side walls of the standard finray-effect gripper bend by applying force, usually from contact when grasping parts, which results in a deformation of the base and tip towards the applied force \cite{crooks2016fin}. However, the standard V-shaped finray design is not ideal to compensate misalignment for parts grasped at the fingertip, as most movement would be in the middle of the finger. Additionally, using the V-shaped finger with tilted crossbeams results in a stiff fingertip. Additionally, applying a force along the finger results in a rotation of the fingertip and therefore a rotation of the contact plane, which could result in contact loss of the gripped part. Instead, a translational deflection at the fingertip and bending of the ribs are desired to compensate misalignment while maintaining contact with the grasped part.

Finally, a V-shaped fingertip does not allow form-closure features or notches introducing a mechanical stop for the gripped cable to positive lock it inside the gripper. The form-closure combined with a high stiffness value in the assembly direction are necessary to realize high assembly forces. 
% The goal of the fingertips with a notch is to introduce a mechanical stop for the gripped cable to positive lock it inside the gripper. This creates a higher structural stiffness in the assembly direction to successfully insert the connector, and allows for a higher force without in-hand slip. 

% The notch allows form-closure, but comes with a trade-off in limiting the range of cables to be handled. The main reason for this is that a too large notch results in a line contact with the cable wires instead of a robust planar contact with the plug's outer surface. This results in an unstable grip and therefore possible slip. Additionally, if the notch is too big, thus nearly enveloping the cable head, there might not be enough space to insert the connector into the plug. However, if the notch is too small, the contact plane could be too small to achieve a form-fit contact, where almost only point contact is realized at the contact area, resulting in an unstable grip. 

While a high stiffness is desired in the assembly direction, significantly lower stiffness is desired laterally, to compensate misalignment and reduce resulting contact forces. To achieve this, the V-shape finger profile is changed. Instead of the outer walls approaching another towards the tip, the distance remains the same overall, with variations in the form of the fingertips, which is discussed in the following section \ref{subsec:design}. 

% From the following paragraph, extract the monolithic + formschluss aspects

% Regarding the requirements and constraints in Sec. \ref{sec:reqconst}, the monolithic compliant gripper can be built to successfully work in narrow environments. This is done without any control methods besides the collision detection from the UR. In future design iterations, a notched fingertip is used to achieve the necessary $15 N$ of assembly force, by form-fitting the various forms and geometries of plugs, which where demonstrated in Fig. \ref{Cables}, inside of the grip, achieving an effectively higher stiffness in the assembly direction. With the compliance realized by this finray-effect gripper, the tolerances described before can be compensated with a certain deflection of the fingers resulting from external forces from contact between the connectors. Using search strategies, described later in Sec. \ref{sec:repeatability_magazine}, allows the gripper to successfully assemble the parts in case of misalignment. Control of the object's pose inside the grip can be assumed to be negligible when combined with a magazine, as described in Sec. \ref{sec:testenv}.

% As seen in Fig. \ref{bending_v} and Fig. \ref{0deg_v}, 

% visualized in Fig. \ref{0deg_roundtip}. This has shown the best results in early tests, however, variations in the form of the fingertips are tested over the following iterations. 

\subsection{Design Parameters}  \label{subsec:design}
Two important parameters of the finger design are optimized to improve performance.

 \subsubsection{Infill Options}
 The most important design parameters are the infill options to adjust the density and orientation of the ribs in the finger, i.e. the infill direction, given in degrees, and the infill density options, given in percentage, as proposed by \cite{Elgeneidy.2019} and visualized in Fig. \ref{finger_design_rotated}. This affects the bulk stiffness realized by the finger on a gripped part as well as the maximum force that the finger can apply.

\subsubsection{Fingertip Options}
An additional parameter is the form of the fingers which can either be with a rounded top, flat top, notched rounded top, flat angled or notched top with a contact plane, visualized in Fig. \ref{finger_design_rotated}. A notched fingertip is necessary to both allow a form-fit connection between gripper and part and to achieve an optimal deflection motion. The notch can be rotated by a certain degree, corresponding to the mounting angle of the finger, 
% inclination of the mount 
used to achieve a parallel contact plane with the grasped part. The size of the notch depends on the cable to be handled, which could limit the target in developing a gripper able to handle a broad variety of cables with a form-fit connection.
However, introducing a notch limits the range of cables to be handled. The main reason for this is that a too large notch results in a line contact with the cable wires instead of a robust planar contact with the plug's outer surface. This results in an unstable grip and therefore possible slip. Additionally, if the notch is too big, thus nearly enveloping the cable head, there might not be enough space to insert the connector into the plug. However, if the notch is too small, the contact plane could be too small to achieve a form-fit contact, where almost only point contact is realized at the contact area, resulting in an unstable grip. 

The friction of PLA+ and PETG, proves to be insufficient for a stable grasp, which is why additionally an adhesive layer could be considered on the contact plane of the fingertip. Multiple grasping modes are possible here, either aspiring a pinch contact of the wires or the cable head, or a parallel grasp of either. The pinch contact could be varied to either achieve a point or line contact with the corresponding part, or to achieve a planar contact.

To compensate misalignment parallel to the moving direction of the gripper's jaws, structured compliance in the base-y-direction is desired. However, unintentional DoFs, as a rotation about the base-y-axis or the base-x-axis resulting in a change of pose of the grasped part resulting from contact forces, are possible. The coordinate system is visualized in Fig. \ref{subfig:key-a}.

\subsection{Manufacturing Process}
To allow for an easy adaptation of infill density and line directions, the parameters are set directly in the slicer program instead of CAD. Here \textit{Ultimaker Cura} is used, applying a method similar to the method used in \cite{Elgeneidy.2019}.  The materials used in this work are orange PLA+  
% from \textit{Zhuhai Sunlu Industrial Co.,Ltd.} 
% \cite{Sunlu}
and black PETG. 
% from \textit{Verbatim GmbH} 
% \cite{verbatim_PETG}, 
% referred to as PETG.

To achieve an easy adjustment, first, the finger has to be designed as a solid in CAD as proposed by \cite{Elgeneidy.2019}. The part is exported as an .stl - file and loaded into \textit{Cura}. Here the parameters for the gripper can be set. Inspired by \cite{Elgeneidy.2019}, the infill type is set to lines in \textit{Cura}, to achieve the desired rib-structure, the option to connect infill lines is turned off. To allow compliance of the skin of the finger, the wall line count has to be set to one, with a line width close to the nozzle diameter of 0.4 mm, slightly deviating from the recommended 2x nozzle diameter from \cite{Elgeneidy.2019}. However, the line width of the infill is also set to 0.4 mm, hence applying the recommendation of \cite{Elgeneidy.2019}, where a line width close to the nozzle diameter is suggested. The top and bottom layers are removed to fully achieve compliance through the ribs. 
% In the first tests, infill densities of $10\%, 20\%$ and $30\%$ are used, with the direction of the infill lines varying in $10\degree$ steps from $0\degree$ to $40 \degree$. 

\begin{figure}[t]
  \begin{subfigure}{0.55\linewidth}
    \includegraphics[width=\columnwidth]{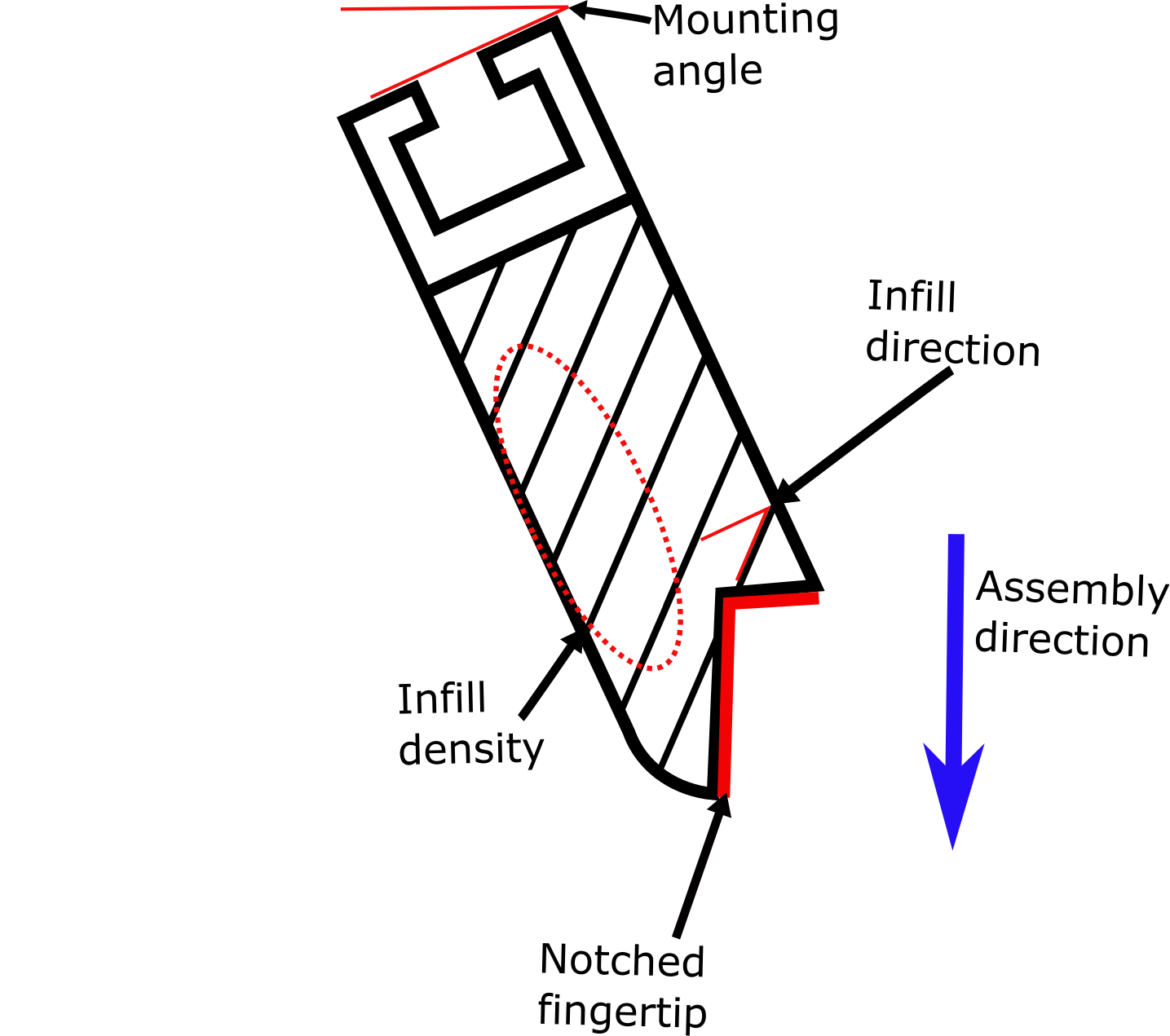}
    \caption{}\label{finger_design_rotated}
  \end{subfigure}\hfill
  \begin{minipage}{0.38\linewidth}
    \begin{subfigure}{\linewidth}
      \includegraphics[width=\linewidth]{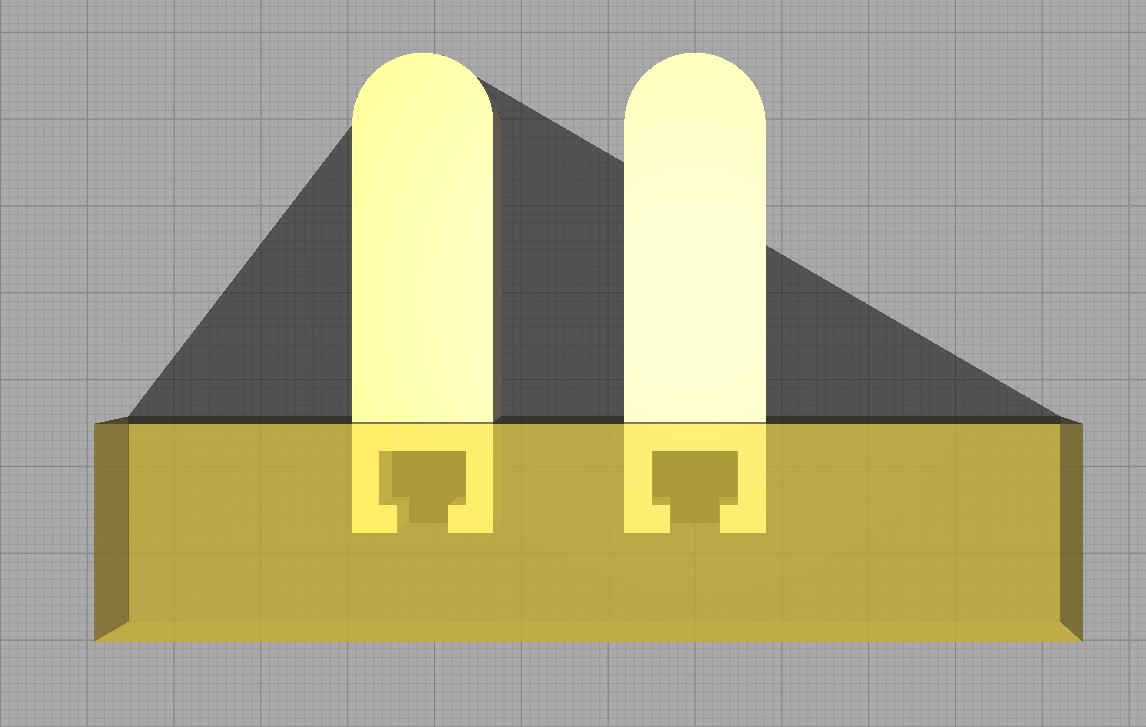}
      \caption{}\label{support_blocker}
    \end{subfigure}\hfill
    \medskip
    \begin{subfigure}{\linewidth}
      \includegraphics[width=\linewidth]{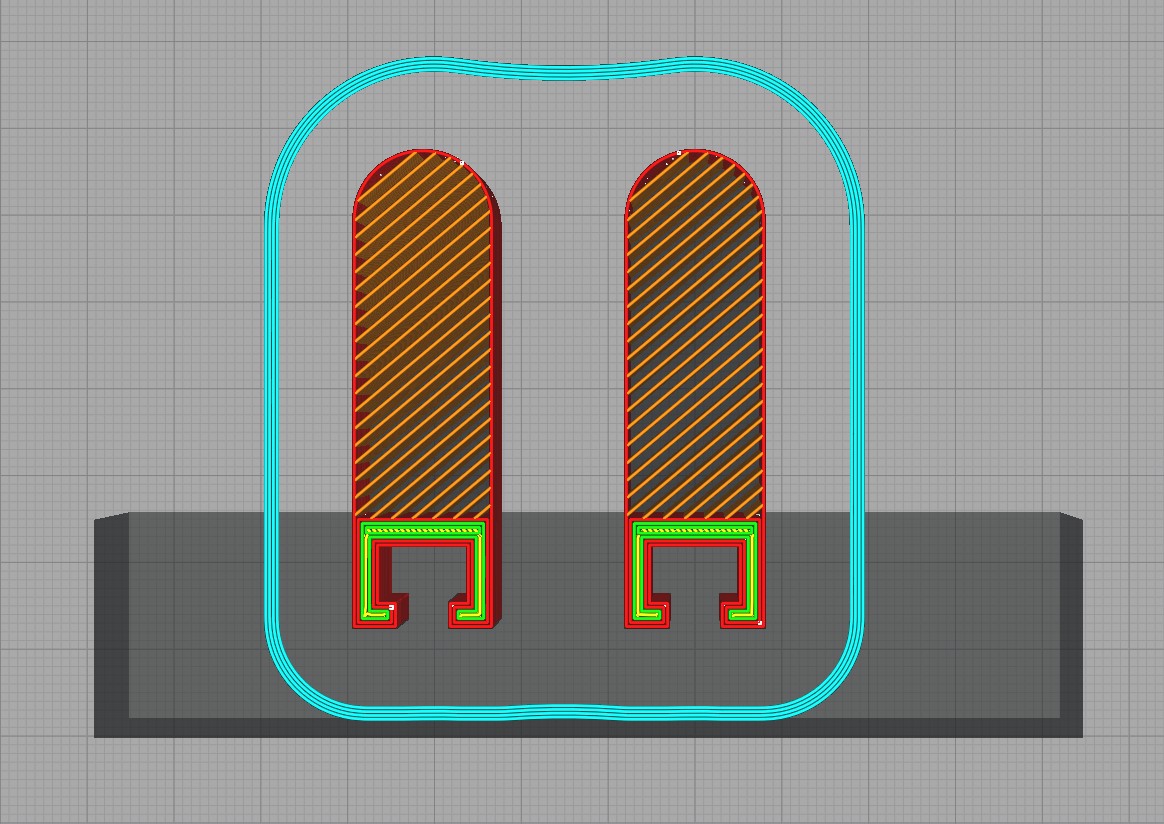}
      \caption{}\label{sliced}
    \end{subfigure}\hfill
  \end{minipage}
  \caption{Design process of the finray-effect gripper, (b) support blockers to allow for varying slicing settings, and (c) sliced gripper with compliant structures at the top and a rigid base}
  \label{fig:design_process}
\end{figure}

However, with these settings the connection between finger and mount to the gripper would also be manufactured the same way, with a high level of compliance and flexibility which is suboptimal for a connection withstanding the applied contact forces of the fingers. For the lower connecting section of the finger different slicing settings have to be used where \textit{Cura's} "support blocker" feature is applied, as visualized in Fig \ref{support_blocker}.
The support blocker allows dividing the two sections of the finger to change selected slicing parameters. For example, other than for the section of the fingers with the ribs, top and bottom layers are needed here. With the option "Per Model Settings" and "Modify Settings for overlaps" the wall/top/bottom thickness, wall line count and top/bottom layers can be changed individually for the section within the support blocker. This allows the part to be manufactured with individual settings, as seen in Fig. \ref{sliced}. The in this paper used .stl/.stp/.ipt files are available at \url{https://github.com/richardhartisch/compliantfinray}.  

% \begin{figure}[t]
%     \centering
%     \subfloat[Support Blocker in \textit{Cura}
% 	\label{support_blocker}]{\includegraphics[width=.50\columnwidth]{figs/finray_support_blocker.jpg}}
% 	\hfill
%     \subfloat[Sliced finrayGripper
% 	\label{sliced}]{\includegraphics[width=.48\columnwidth]{figs/finray_sliced.jpg}}
%     \caption{Design process of the finray-effect gripper, (a) support blockers to allow for varying slicing settings, and (b) sliced gripper with compliant structures at the top and a rigid base}
%     \label{flexures}
% \end{figure}

\section{Validation}
\label{sec:app}

This section gives an overview of the process used to iteratively test and validate the fingers for the assembly of a plug into socket.   
%To further optimize the search process and to optimally compare the different parameters of the fingers, failures due to the plug and sockets attributes, which where described before in Sec. \ref{subsec:failures}, have to be minimized as far as possible. 
% Hence, for the following experiments and to allow tests in an industrial application, an "automation-friendly" plug and socket connection is used. The test environment is equipped with the spring and magazine introduced in Sec. \ref{sec:testenv}. 
The goal is to successfully pick the plug from a magazine and assemble into a socket with and without various misalignment values. 
% As described, the movement of the cable head inside the grip can also lead to failure. This is caused by the weight of the cable exceeding the maximum applicable force of friction consisting of the gripping/normal force and the coefficient of friction given by the adhesive layer. The weight is composed of the weight of the cable head and the wires, as seen in Fig. \ref{automat_cable}. 
The programs used are programmed via the teach panel of the \textit{Universal Robots UR5}. The program is intended for high-speed assembly, with tool speed values of 250, up to 700 $mm/s$ and a tool acceleration of 1200, up to 2000 $mm/s^2$, using the MoveL command of the \textit{UR} to approach the waypoints.

The assembly and grasping process is summarized as follows. First, the cable is grabbed from the magazine. The contact force of the fingers overcome the contact force of the spring when moving the gripper in a linear movement upwards, with which the cable is removed from the magazine. To compensate slippage during the first phase, afterwards the gripper could push the cable head slightly on the table with a linear movement downwards to ensure a contact with the upper contact surface of the finger. Now, in the second phase, assembly takes place, using the search strategy described earlier in \ref{subsec:strategies}. A video is available at \url{https://youtu.be/J7EGXtE54oY}. 

%Generally speaking, the movement values depend on the present task. 
% The connector is held angled in the initial position. The first contact consists of a planar contact of one side of the connector and the front part of the housing of the socket. The first contact point is successful when the front ridge of the connector lies flat on the front part of the housing. The next contact point is achieved by sliding the connector in the negative base-x-axis while moving slightly in the negative base-z-direction. With this, the connector is pushed slightly into the socket while a contact with the back wall is aspired. The connector should now slightly be inserted into the plug. However, there still is no contact with the front wall and at least one side wall to fully align the connector according to the constraints. For this, in the next step, the connector is moved in the positive base-x-direction to achieve contact with the inside of the front wall. To fulfill the last condition, the plug is moved in the positive base-y-direction, so that the side of the connector is in contact with the inside of the socket's side wall.
% Now, since all constraints have been complied with, for the final step, the plug is moved linearly in the negative base-z-direction to assemble the connectors, as seen in Fig. \ref{exp_3_fifth}. 

% After a successful assembly, the closed gripper rotates around the x-axis, releases the grip and moves back to the initial position. To reset, the connector can now be removed from the plug and is clamped back into the magazine.

\subsection{Repeatability Experiments}
\label{sec:repeatability_magazine}

To test for repeatability, the assembly process is repeated 84 times with a fixed socket position, using the aforementioned UR program and manually resetting the plug in the magazine, out of which the assembly failed twice. The first failure occurred at attempt 30 and the second failure at attempt 84 which ultimately lead to a component failure of the fingers. This concludes a roughly 97,6 \% success rate. 
Assumed causes are either slight slippage in the grip coming from the adhesive tape, or the kinematics of the robot. Additionally, the table on which the robot is mounted is not fixed but on wheels, which could add another level of instability, impairing the robustness. 

\subsection{Robustness Experiments}
\label{sec:robustness_exp}

In the next experiments, the robustness over variation in socket position is tested, to clarify the impact of the design parameters on the robustness and tolerable range. For the initial test, the boundaries of compensable misalignment of the plug to the socket are determined in x- and y-direction in $0.5 mm$ steps. To control the misalignment, instead of a fixed waypoint for the socket position, a variable waypoint is programmed which can be changed in each iteration.  A finger with 0° infill direction, 10\% infill and a 10° mount are used. A successful assembly is repeated five times to assure repeatability. If five assemblies in a row are successful, another 0.5 mm is added to the misalignment and the sequence of five trials starts again. This is repeated until the maximum compensable misalignment is met and the assembly fails for the first time. This is done to test the limits both in the  x- and y-direction. With this setup, it can be shown that with a 100\% speed value of the program and the used search algorithm this compliant finger design is capable of tolerating a misalignment in a range of 7.5 mm in y- direction and 7 mm in the x- direction. 
To further compare the tolerance windows with varying designs, the limits of the first run are tested with varying infill densities and infill directions. The results are listed in Table \ref{tab:erg_robustness}. 
% Here, $m$ denotes mount, meaning which mount configuration is used (either 10° and 20° have been tested), $i$ denotes the infill density in percentage and $id$ is an abbreviation for the infill direction in deg.  

\begin{table}[ht]
  \centering
  \caption{Results Robustness Experiment, $m$ denotes mount, meaning which mount configuration is used (either 10° and 20°), $i$ denotes the infill density in percentage and $id$ is an abbreviation for the infill direction in deg. }
  \label{tab:erg_robustness}

     \begin{adjustbox}{width=0.95\textwidth/2}

    \begin{tabular}{|c|c|c|c|c|c|c|c|c|c|c|}
 & x & y & x & y & y & y & y & y & y & y\\
 & 10° m & 10° m  & 20° m & 20° m & 10° m & 10° m & 10° m & 10° m & 10° m & 10° m  \\

Variant & 10\% i & 10\% i & 10\% i & 10 \% i & 20\% i & 30\% i & 10\% i & 20\% i & 30\% i  & 10\% i \\

 & 0° id & 0° id  & 0° id & 0° id & 0° id & 0° id & 0° id & 0° id & 0° id & 10° id   \\
 & PLA+ & PLA+ & PLA+ & PLA+ & PLA+ & PLA+ & PETG & PETG & PETG & PLA+\\

\hline
\hline

range [mm] & 7 & 5.5 & 4.5 & 5.5 & 5 & 4.5 & 7.5 & 6 & 5.5 & 5.5  \\

\hline 

    \end{tabular}

  \end{adjustbox}

  \vspace{5pt}
    \begin{adjustbox}{width=0.95\textwidth/2}
    \begin{tabular}{|c|c|c|c|c|c|c|c|c|c|c|c|c|}
 & y & y & y & y & y & y & y & y & y & y & y \\
 & 10° m & 10° m & 10° m & 10° m & 10° m & 10° m & 10° m & 10° m & 10° m & 10° m & 10° m \\

Variant& 10 \% i & 10 \% i& 15 \% i & 20 \% i & 25 \% i& 30 \% i & 10 \% i & 15 \% i & 20 \% i & 25 \% i& 30 \% i  \\

 & 20° id & 30° id & 30° id & 30° id & 30° id & 30° id & 40° id & 40° id & 40° id & 40° id & 40° id  \\
 & PLA+ & PLA+ & PLA+ & PLA+ & PLA+ & PLA+ & PLA+ & PLA+ & PLA+ & PLA+ & PLA+ \\

\hline
\hline

range [mm] & f & f & 5.5 & 5.5 & 5.5 & 6 & f & f & f & 2 & 5.5  \\

\hline

    \end{tabular}    
 \end{adjustbox}

\end{table}

It is important to note that regarding the compensable range in x-direction, the compensation is attributable to the free rotation of the cable head inside the grip about the base-y-axis, corresponding to the coordinate system visualized in Fig. \ref{subfig:key-a}, and should be treated as a positive side-effect of the finger's design, which can be described as an unforeseen DoF. 
% This is also another reason as to why the contact plane in the notched fingertip is rejected.
The main focus of the finger's design is to allow a compliance in the y-direction due to the ribs, which is why the experiment only shows general feasibility of a compensation in the x-direction for the 10° and 20° mounts but does not compare the tolerance in x-direction for every finger, as seen in Tab. \ref{tab:erg_robustness}.

\subsection{Discussion}
As Tab. \ref{tab:erg_robustness} shows, with a 10° mount the tolerable misalignment-range is slightly bigger than with a 20° mount.  During the tests for 20° infill direction and 10 \% infill density at a misalignment of $5 mm$ early signs of buckling are noticed. At the next increment of the infill direction, this is noticed already at $4 mm$ and plastic deformations at the connections of the ribs to the outer wall appear at $4.5 mm$. 30° infill direction with 10 \% infill initially stands out due to a comparably large compensable range of $\approx 5.5 - 6.5 mm$. However, beginning with $+4 mm$ misalignment, some slight buckling can be observed building up to slight plastic deformation at the connection of the ribs to the outer wall with the next misalignment increments, which is why this variant is not considered ideal.  
The extreme value of 40° direction shows to be difficult to test. At a comparably low misalignment value of $2 mm$, component failure already occurs for 10\% infill density resulting in a non-feasible combination for any assembly tasks. This is also noticeable for 15\% infill density where buckling and component failure occurred at $3 mm $ and $3.5  mm$. Because the initial start-value is set too high, the part is already permanently damaged, resulting in a failed assembly at $-1.5 mm$ and $-2 mm$. At 20 \% infill there is strong bulging noticeable at $2 mm$ and buckling at $2.5 mm$. $-2 mm $ proves to be compensable, however, bulging is noticeable here, too. With 25 \% infill strong deformation is noticed at $3.5 mm$. $4 mm$ is also successful, however, some plastic deformation occurs, which is why the experiment is stopped here to prevent any further damage. The last increment for the infill density at 40° infill direction proves to be the most stable one. Some strong deformation is observed at $3 mm$ but without plastic deformation. At $-3 mm$ the cable head strongly clips into the plug, which is why no further tests are done for this variant to prevent any further damage. This is attributable to an excessive vertical stiffness of the finger, where compliance is still present, with a potentially too high contact force profile which could damage the electrical components. Thus, this variant should not be used to assemble delicate parts.   

% The robot's collision detection by the \textit{UR} proves to be successful as well. During the tests for 0° infill direction and 20 \% infill density and a misalignment of $- 2 mm$, where a failure is anticipated, the protective stop engages before damaging the fingers. 

Regarding the PETG fingers, 10° infill direction and 10\% infill density proves to achieve the biggest tolerance range of all combinations, tolerating $\approx 7.5 mm$ misalignment. However, this combination is not suitable for any assembly tasks because the cable head slips easily inside the grip. This is traced to a very low gripping force from the fingers due to the infill settings. Increasing the infill density by 10 \% already results in a better grip, while achieving a tolerance range of $\approx 6 mm$. Another 10 \% show similar results, the compensation of $+3.5 mm$ misalignment cannot be repeated robustly. Using PETG comes with the benefit of a higher flexibility compared to PLA+, which results in a lower risk of plastic deformation during handling. 
The tolerable range can be defined as $\approx 5.5 mm$ while providing a stable grasp on the gripped cable head.

\section{Conclusion and Future Work}

\label{sec:discussion}
To the authors' best knowledge, this work has proposed a first use of a finray-effect gripper for structured compliance. Other than previous works using the finray-principle, which focus on a stable grasp on objects with varying surface geometry, this design is here realizes directionally-dependent stiffness on the gripped part. This used to robustly and repeatedly compensate misalignment  in the range of up to $7.5$ mm in high-speed assembly tasks. Additionally, the objective, as defined before in Sec. \ref{sec:intro}, of achieving a comparable success time as in \cite{park2013intuitive} is reached and exceeded, as the assembly time from  first contact is $\approx$ 1.2s. Hence, feasibility of the passive compliant fingers to compensate misalignment in high-speed tasks without additional sensing is proven. 

For an optimal finger design, the finger stiffness, the maximum tolerable force, the maximum deflection, the gripping stability and the compensable range have to be taken into consideration. A variant with a too high stiffness, e.g. variants with a 30\% infill density, especially with an increasing infill direction could damage the assembly parts. A too low stiffness, e.g. PETG with 10 \% infill density would not be able to lift and transport the cable robustly and maintain a stable grip when external forces occur. Choosing a 40° infill direction results in component failure due to plastic deformation for almost every variant.
Most of the variants listed in Tab. \ref{tab:erg_robustness} achieve a tolerable range of $\approx 5.5 mm$, 30 \% infill density and 0° infill direction achieves the lowest, with $\approx 4.5 mm$. PETG shows the best results here, with a maximum range of $\approx 7.5 mm$ for the non-applicable 10 \% infill variant. Thus, the higher rib angle PETG variants are recommended in this case. 

% The compensation of combined misalignment values has not been studied in this work. This is because only the passive compensation structures as the controllable parameters could be adjusted, the compensation in the x-direction are primarily unintentional and a result of a free rotational movement inside the grip. Hence, an investigation of the combined values would not provide clear conclusions of the density and direction of the fingers' ribs.

Future work will focus on determining the mechanical attributes regarding the stiffness, max tolerable force and hysteresis experimentally. With this, additional attempts to design the fingers by using FEA or by analytically determining the mechanical properties and to achieve a better intuition of how the design parameters influence the final stiffness of the structure. 

Using fused deposition modeling as an additive manufacturing process comes with its own limitations, as the direction in which the part is built up has to be considered. Certain structures need an optimal orientation to the print bed to be successfully manufactured, as overhangs or otherwise unsupported structures could fail without support. Using alternative manufacturing processes could allow one to create ribs in varying directions which could introduce multi-directional structured compliance into the finger. Additionally, other material could be used which could achieve higher contact forces and would be less sensitive to wear and fatigue.  

\bibliographystyle{IEEEtran}
\bibliography{lib2.bib} % Export for bibtex! Not biblatex. 
\end{document}